\documentclass[conference]{IEEEtran}

\usepackage{cite}
\usepackage{amsmath,amssymb}
\usepackage{graphicx}
\usepackage{booktabs}
\usepackage{array}
\usepackage{url}
\usepackage[T1]{fontenc}

\graphicspath{{./}}

\author{
\IEEEauthorblockN{Xiao Han}
\IEEEauthorblockA{\textit{Goizueta Business School}\\
\textit{Emory University}\\
Atlanta, GA, USA\\
xhan@alumni.emory.edu}
\and
\IEEEauthorblockN{Jingjing (May) Liu}
\IEEEauthorblockA{\textit{Electrical Engineering and Computer Sciences}\\
\textit{University of California, Berkeley}\\
Kirkland, WA\\
mayliujj@berkeley.edu}
\and
\IEEEauthorblockN{Moxuan Zheng}
\IEEEauthorblockA{\textit{Stern School of Business}\\
\textit{New York University}\\
New York, NY, USA\\
mz2156@nyu.edu}
\and
\IEEEauthorblockN{Zhen Zhang}
\IEEEauthorblockA{\textit{School of Data Science}\\
\textit{University of Pennsylvania}\\
Philadelphia, PA, USA\\
billzhangzhen98@gmail.com}
\and
\IEEEauthorblockN{Chenyu Wu}
\IEEEauthorblockA{\textit{Pratt School of Engineering}\\
\textit{Duke University}\\
Durham, NC, USA\\
wuchenyu999@outlook.com}
}

\title{Interpretable versus Learned Encoders for High-Cardinality Fraud Detection}

\begin{document}

\maketitle

\begin{abstract}
A total of seven categorical encoding methods were tested on the IEEE-CIS fraud benchmark dataset (590{,}540 records, 3.5\% positives, 8 high-cardinality columns). The encoders were evaluated using a stratified 5-fold cross-validation (CV) with three repetitions. Five of the encoders had identical frozen LightGBM learners in the downstream phase, allowing for controlled comparisons of their performance to each other. CatBoost and TabNet were included as comparisons across paradigms using different learners. The entity embeddings produced the highest AUC-ROC (\textbf{0.9612}), with a statistically significant tie with that of CatBoost (0.9602) and statistically superior to tier group encoding (0.9548), whereas target encoding was only 0.0023 worse than tier group encoding and the auditor-friendly tier boundaries were maintained. Off-the-shelf TabNet did not outperform tree-based pipelines and collapsed under data scarcity. On AUC-PR, CatBoost leads (0.822 vs.\ 0.793); no encoder dominated both metrics. Per-column analysis confirmed the embedding advantage arises from joint multi-column representation.
\end{abstract}


\section{Introduction}

Tabular transaction records contain many high-cardinality categorical variables (card identifiers, billing addresses, email domains, device fingerprints) with cardinalities from four to tens of thousands. How to represent such categories is an important modeling decision, but evidence is inconsistent: encoder benchmarks stay within the classical family~\cite{pargent2022regularized,cerda2020encoding}, while tabular deep-learning studies fix the encoding and vary the model~\cite{grinsztajn2022treebased,shwartzziv2022tabular,borisov2022survey}. A banking practitioner cannot determine whether neural entity embeddings~\cite{guo2016entity} or a leakage-safe target encoder~\cite{miccibarreca2001target,prokhorenkova2018catboost} will serve best in a fraud model when model-transparency requirements (such as SR~11-7 in the US, or analogous audit standards elsewhere) demand that scores be explainable.

The question spans three axes. \emph{Accuracy} margins are narrow, so isolating the encoder from the learner matters. \emph{Compute} varies widely: end-to-end deep models can be an order of magnitude slower than tree-plus-embedding pipelines. \emph{Interpretability} is critical under SR~11-7, which requires model-risk owners to justify scores, so a slightly less accurate but more interpretable encoder may be preferable. We report all three together.

We address: \emph{what are the accuracy, interpretability, and computational trade-offs of seven encoders on a large fraud dataset with a fixed downstream learner, and where does lightweight statistical grouping sit on this continuum?} Our contributions:

\begin{enumerate}
\item \textbf{Cross-paradigm controlled experiment.} We compared five encoders under a single fixed LightGBM learner on the IEEE-CIS dataset with stratified CV, plus CatBoost and TabNet as cross-paradigm reference points. Previous studies~\cite{grinsztajn2022treebased,shwartzziv2022tabular} varied the model family with a default encoder; we varied only the encoder, finding $\approx$0.01 AUC-ROC attributable to encoding alone.
\item \textbf{Auditor-readable encoding operationalized and benchmarked.} We operationalized and benchmarked target-aware tier grouping (Bayesian-smoothed fraud rates binned into $K$ ordinal tiers~\cite{miccibarreca2001target,slakey2019bayesian}) and evaluated it against learned representations.
\item \textbf{Interpretability-first framing for regulated finance.} We scored interpretability, compute, and metric robustness as primary outcome dimensions alongside SR~11-7 model-risk-management expectations.
\end{enumerate}

\section{Related Work}

\textbf{Categorical encoding.} Target encoding with Bayesian smoothing~\cite{miccibarreca2001target} underpins leakage-safe variants in CatBoost~\cite{prokhorenkova2018catboost}. Other methods include feature hashing~\cite{weinberger2009hashing}, similarity encoding~\cite{cerda2018similarity,cerda2020encoding}, and conjugate Bayesian encoders~\cite{slakey2019bayesian}. Pargent et~al.~\cite{pargent2022regularized} find regularized target encoding best among classical encoders; Guo and Berkhahn~\cite{guo2016entity} established entity embeddings as the deep-learning default. Our tier grouping applies the smoothing principles of~\cite{miccibarreca2001target,slakey2019bayesian} and bins smoothed rates into ordinal tiers; we evaluate it as an auditor-readable encoder.

\textbf{Trees versus deep learning on tabular data.} Recent deep models include TabNet~\cite{arik2021tabnet}, NODE~\cite{popov2020node}, and FT-Transformer~\cite{gorishniy2021fttransformer}. Grinsztajn et~al.~\cite{grinsztajn2022treebased} and Shwartz-Ziv and Armon~\cite{shwartzziv2022tabular} show GBDTs (XGBoost~\cite{chen2016xgboost}, LightGBM~\cite{ke2017lightgbm}) still match or exceed them; Borisov et~al.~\cite{borisov2022survey} identify interpretability and compute as unresolved trade-offs. None of these studies compare encoders across both paradigms.

\textbf{Closing the divide.} Encoder benchmarks use classical learners~\cite{pargent2022regularized,cerda2020encoding}; tree-vs-deep studies fix the encoding~\cite{grinsztajn2022treebased,shwartzziv2022tabular}. No study has treated the encoder as the controlled variable across both. We close this gap on IEEE-CIS~\cite{ieee2019fraud}, a 590K-transaction fraud task ($\approx$3.5\% base rate)~\cite{nguyen2020deeplearning,randhawa2018payment,fiore2019credit,sun2025highrecall}, keeping the learner, folds, and protocol constant~\cite{caruana2006empirical}.

\section{Methods}

We defined seven encoders on the IEEE-CIS fraud dataset (590{,}540 records, 3.50\% positive, eight high-cardinality columns, cardinalities 4--13{,}553). E1--E4 and E6 feed the same downstream LightGBM~\cite{ke2017lightgbm} with frozen hyperparameters; E5 uses CatBoost~\cite{prokhorenkova2018catboost} as both encoder and learner; E7 feeds target-encoded input to TabNet~\cite{arik2021tabnet}. All encoders receive identical raw features (eight high-card columns, eleven low-cardinality one-hot columns, 395 numerical columns). We partition results into \emph{controlled comparisons} (E1--E4, E6: same learner) and \emph{cross-paradigm observations} (E5, E7: different learner).

The eight high-cardinality target columns (with cardinalities) are: \texttt{card1} (13{,}553), \texttt{DeviceInfo} (1{,}786), \texttt{addr1} (332), \texttt{addr2} (74), \texttt{R\_emaildomain} (60), \texttt{P\_emaildomain} (59), \texttt{card4} (4), \texttt{card6} (4).

\textbf{E1: One-hot.} Each value becomes an indicator; values that occurred fewer than fifty times in the training fold are dropped ($\approx$2{,}912 features per fold).

\textbf{E2: Target encoding.} Each value is replaced with its out-of-fold mean target via five-fold inner cross-fitting with Bayesian smoothing ($m{=}30$); a permutation test confirms leakage invariance.

\textbf{E3: Frequency.} Each value maps to its relative frequency in the training fold.

\textbf{E4: Tier grouping (interpretable).} Per high-cardinality column on each fold's training partition: (1)~compute Bayesian-smoothed fraud rate $\hat{r}_v = (n_v r_v + m\bar{r})/(n_v + m)$, $m{=}30$; (2)~sort by $\hat{r}_v$ and bin into $K \in \{3,5,7\}$ ordinal tiers via equal-frequency, KS-greedy, or chi-square binning; (3)~replace raw values with tier~IDs, monotone in fraud rate. Unseen test values map to the tier closest to $\bar{r}$. Tiers are auditor-readable: an inspector sees ``tier~5 $=$ outlook.com at 16.5\%, tier~1 $=$ comcast at 1.2\%'' from bin boundaries. Because several columns are skewed (\texttt{addr2} is 88\% one value), $K$ is an upper bound on populated tiers.

\textbf{E5: CatBoost.} CatBoost~\cite{prokhorenkova2018catboost} consumes raw categoricals natively via a permutation-based ordered target statistic, avoiding leakage without explicit inner CV. Hyperparameters mirror LightGBM (\texttt{iterations=500}, \texttt{learning\_rate=0.05}, \texttt{depth=7}, \texttt{early\_stopping\_rounds=50}).

\textbf{E6: Entity embeddings $\to$ LightGBM.} Following Guo and Berkhahn~\cite{guo2016entity}, a shallow network learns dense embeddings per high-card column; trained embeddings are extracted and fed to the same frozen LightGBM. Per-column dimension is $\min(50, (n_\text{unique}+1)//2)$ (255 total). Classifier head: two Linear--ReLU--Dropout(0.3) blocks (256, 128), Adam ($lr{=}10^{-3}$), batch~4096, 20 epochs, patience~3. The extracted matrix averages $\approx$1{,}824 columns. For each outer fold, the embedding network is fit \textbf{only} on the training partition; validation labels enter solely via early-stopping. TabNet (E7) follows the same rule.

\textbf{E7: Target encoding $\to$ TabNet.} TabNet~\cite{arik2021tabnet} with published defaults (\texttt{n\_steps=3}, \texttt{n\_a=n\_d=64}, $\gamma{=}1.5$, batch~4096, 50 epochs, patience~5). E7 receives the same pre-encoded numerical feature matrix as E2 (low-cardinality columns one-hot, high-cardinality columns target-encoded) and does not consume raw categorical indices; it differs from E2 only in the downstream learner (TabNet vs.\ LightGBM), so the E7--E2 contrast isolates the learner with encoding held fixed. No architecture or hyperparameter search was performed.

\section{Experimental Setup}

\textbf{Dataset.} IEEE-CIS Fraud Detection~\cite{ieee2019fraud}: 590{,}540 transactions, 3.50\% positive, 432 features. Eight high-cardinality columns (\texttt{card1}, \texttt{DeviceInfo}, \texttt{addr1}, \texttt{addr2}, \texttt{R\_emaildomain}, \texttt{P\_emaildomain}, \texttt{card4}, \texttt{card6}) are the encoding targets; eleven low-cardinality columns are always one-hot encoded identically.

\textbf{Protocol.} Stratified 5-fold CV $\times$ 3 repeats $=$ 15 runs per encoding. LightGBM hyperparameters fixed across E1--E4, E6 (\texttt{n\_estimators=500}, \texttt{learning\_rate=0.05}, \texttt{max\_depth=7}, \texttt{early\_stopping\_rounds=50}), with no per-encoding tuning. The robustness experiment retrains all encoders on 50/25/10\% subsamples (5 folds per cell, 105 runs); the per-column experiment trains E2, E4, E6 on one high-card column at a time (5 folds per cell, 120 runs).

\textbf{Metrics.} We fixed \textbf{AUC-ROC as the primary metric} in advance of running the experiments and report \textbf{AUC-PR as a secondary, imbalance-aware metric}. Fixing the primary metric beforehand blocks ex-post metric-shopping; we keep AUC-ROC primary precisely because the two metrics disagree on the winner (Section~V-D).

\textbf{Statistical tests.} Our $R{=}15$ resamplings have correlated folds, so ordinary paired tests understate variance. We take the \textbf{Nadeau--Bengio corrected resampled $t$-test}~\cite{nadeau2003inference} ($n_\text{test}/n_\text{train}{=}0.25$, df${=}14$) as primary pairwise inference. For an illustrative omnibus ranking we run Friedman + Nemenyi~\cite{demsar2006statcomparisons} ($\chi^2{=}80.9$, $p{=}2.3\times10^{-15}$, CD$_{0.05}{=}2.325$); the CD diagram serves as a ranking aid alongside the NB-corrected test.

\textbf{Hardware and reproducibility.} All encoders ran on the same Apple-Silicon machine (neural models on MPS GPU, tree models on CPU). MPS is low-throughput relative to NVIDIA, so absolute wall-clock times overstate DL cost; the cost \emph{ranking} E4$<$E6$<$E5$<$E7 is architectural. Seeds: fold splits \{42,43,44\}, models 42. Software: LightGBM~4.6, CatBoost~1.2.10, PyTorch~2.12.1, pytorch-tabnet~4.1.0. The IEEE-CIS dataset is public~\cite{ieee2019fraud}; all hyperparameters and splits are in Sections~III--IV. Reproducibility of execution assumptions is an active concern in financial ML~\cite{yao2026agentarchitecture}.

\section{Results}

\subsection{Main results and significance}

Table~\ref{tab:main} reports mean $\pm$ std over 15 runs, sorted by AUC-ROC. Scores span only 0.9515--0.9612 ($\approx$0.97\,pp), with entity embeddings (E6) leading at \textbf{0.9612}, narrowly ahead of CatBoost (E5, 0.9602).

\begin{table}[!t]
\renewcommand{\arraystretch}{1.0}
\setlength{\tabcolsep}{4pt}
\caption{Main results, 15 runs (stratified 5-fold CV $\times$ 3 repeats). Sorted by AUC-ROC. Type: D$=$Deep, C$=$Classical, ML$=$ML-native, ip$=$interpretable.}
\label{tab:main}
\centering
\scriptsize
\begin{tabular}{@{}lp{1.8cm}rrrr@{}}
\toprule
ID & Encoding & AUC-ROC & AUC-PR & Fit/s & \#Feat \\
\midrule
\textbf{E6} & Emb.$\to$LGBM (D) & \textbf{0.9612\,$\pm$\,.0020} & 0.7928\,$\pm$\,.0062 & 104.6 & 1824 \\
E5 & CatBoost\,(ML) & 0.9602\,$\pm$\,.0017 & \textbf{0.8216\,$\pm$\,.0058} & 565.3 & 811 \\
E2 & Target\,(C) & 0.9571\,$\pm$\,.0023 & 0.7797\,$\pm$\,.0046 & 79.2 & 1579 \\
\textbf{E4} & Tier grp.\,(C,ip) & 0.9548\,$\pm$\,.0025 & 0.7704\,$\pm$\,.0055 & 24.5 & 1579 \\
E7 & Tgt$\to$TabNet\,(D) & 0.9522\,$\pm$\,.0030 & 0.7828\,$\pm$\,.0111 & 1654.3 & 1579 \\
E3 & Frequency\,(C) & 0.9518\,$\pm$\,.0027 & 0.7631\,$\pm$\,.0063 & 33.9 & 1579 \\
E1 & One-hot\,(C) & 0.9515\,$\pm$\,.0024 & 0.7610\,$\pm$\,.0058 & 24.8 & 2912 \\
\bottomrule
\end{tabular}
\end{table}

Table~\ref{tab:nb} provides the NB-corrected test results that settle the anchor contrasts. E6 significantly outperforms both E4 ($+0.0064$, $t{=}11.74$, $p{<}0.001$) and E7 ($+0.0091$, $t{=}6.95$, $p{<}0.001$). Conversely, E4 vs.\ E7 ($p{=}0.089$) and E6 vs.\ E5 ($p{=}0.23$) are statistical ties. It is also noteworthy that \textbf{E2 outperforms E4} ($-0.0023$, $p{=}0.0003$), meaning tier grouping is competitive with but slightly below the classical champion.

\begin{table}[!t]
\renewcommand{\arraystretch}{1.0}
\setlength{\tabcolsep}{3pt}
\caption{Nadeau--Bengio corrected resampled $t$-test (correlated CV folds, $R{=}15$, $n_\text{test}/n_\text{train}{=}0.25$, df${=}14$). Positive $\Delta$ $=$ first encoder ahead.}
\label{tab:nb}
\centering
\scriptsize
\begin{tabular}{@{}lrrrll@{}}
\toprule
Contrast & $\Delta$ & $t$ & $p$ (NB) & Verdict \\
\midrule
E6--E4 (emb.\,vs\,tiers)    & $+0.0064$ & $11.74$          & $<0.001$ & sig.\,(E6) \\
E6--E7 (emb.\,vs\,TabNet)   & $+0.0091$ & $\phantom{0}6.95$ & $<0.001$ & sig.\,(E6) \\
E4--E2 (tiers\,vs\,target)  & $-0.0023$ & $-4.79$           & $0.0003$ & sig.\,(E2) \\
E4--E7 (tiers\,vs\,TabNet)  & $+0.0027$ & $\phantom{0}1.83$ & $0.089$  & tie \\
E6--E5 (emb.\,vs\,CatBoost) & $+0.0010$ & $\phantom{0}1.26$ & $0.23$   & tie \\
\bottomrule
\end{tabular}
\end{table}

\begin{figure}[!t]
\centering
\includegraphics[width=\columnwidth]{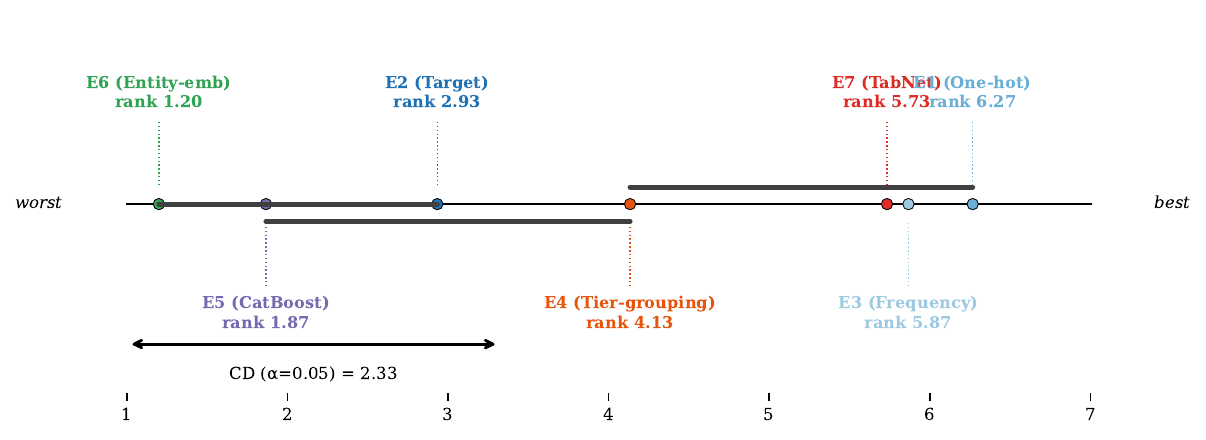}
\caption{Per-run ($N{=}15$) Nemenyi critical-difference diagram (illustrative; CV folds are correlated). Encoders not joined by a bold bar differ at $\alpha{=}0.05$ (CD $=$ 2.325). Primary inference is the NB-corrected test (Table~\ref{tab:nb}).}
\label{fig:cd}
\end{figure}

The Friedman/Nemenyi CD diagram (Figure~\ref{fig:cd}) is illustrative only, since CV folds violate the cross-dataset independence the test assumes. The fold-level CD ($N{=}5$, $k{=}7$) does not account for repeat-level pairing, limiting its power; it reads E6--E4 as a tie ($\Delta$rank 3.00 vs.\ CD${=}4.03$) while the properly powered NB-corrected test finds significance. The cliques are consistent with Table~\ref{tab:nb}: E6/E5/E2 in the top band, E4 just below (E2 statistically ahead at $p{=}0.0003$), and E7/E3/E1 in the bottom band with E7 indistinguishable from E4 ($p{=}0.089$).

\subsection{Accuracy--compute frontier}

Compute cost differences are much larger than accuracy differences. With tier grouping (E4) as the cheapest encoder (24.5\,s/fold, clearing 0.954 AUC), embeddings (E6) gain $+0.0064$ AUC at \textbf{4.3$\times$} the time (104.6\,s). TabNet (E7) costs approximately \textbf{67$\times$} more (1654.3\,s/fold) yet scores 0.0091 below E6. E7's fit time varies (1263--2255\,s) due to the sample-size sensitivity of the sparse-attention architecture; our comparisons use default published parameters with no architecture or hyperparameter search. Given the total AUC-ROC spread is $\approx$0.97\,pp, practical encoder selection depends on compute cost, interpretability, and metric choice.

\subsection{Interpretability and what the embeddings encode}

\begin{figure}[!t]
\centering
\includegraphics[width=\columnwidth]{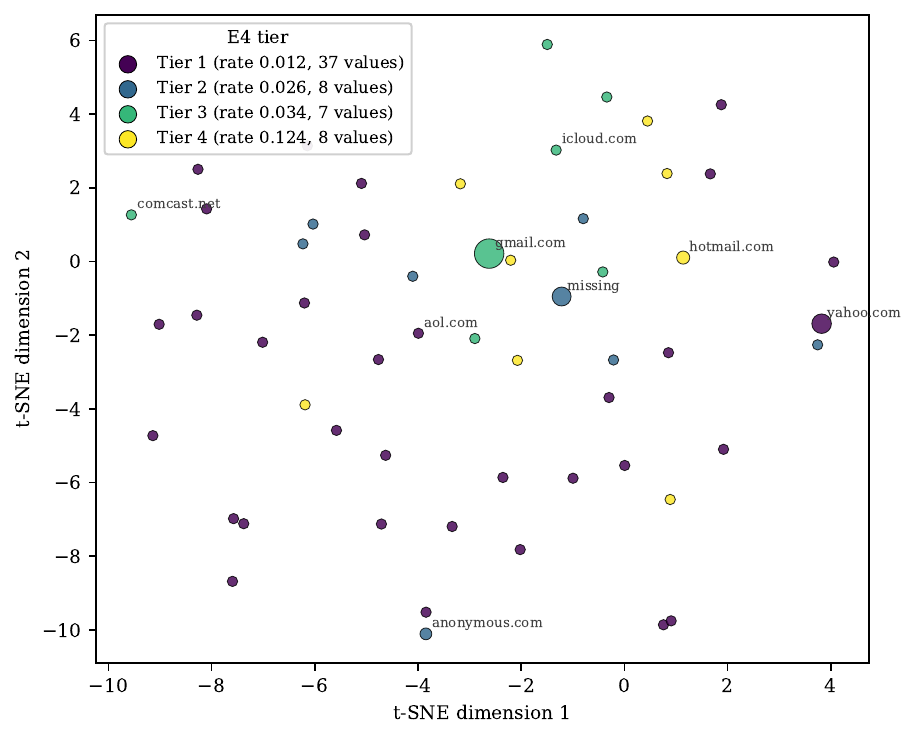}
\caption{t-SNE of E6's \texttt{P\_emaildomain} embeddings, colored by E4 tier. The weak spatial alignment (ARI\,$=$\,0.051) shows embeddings capture structure beyond single-column fraud-rate ordering.}
\label{fig:tsne}
\end{figure}

Figure~\ref{fig:tsne} projects E6's \texttt{P\_emaildomain} embeddings (30-dim) to 2-D via t-SNE, colored by E4's tiers. K-means on the 2-D projection vs.\ E4's tier assignment gives ARI \textbf{0.051} and NMI \textbf{0.175}; because t-SNE does not preserve distances, these values are illustrative of weak visual alignment only and do not quantify representational similarity. E4's tiers encode single-column fraud-rate ordering; the per-column probe (Section~V-E) provides the sound evidence that E6's embeddings capture cross-column structure the tiers discard.

\subsection{AUC-ROC vs AUC-PR: the metrics disagree}

Because fraud detection is highly imbalanced (3.5\% positive), AUC-PR is the more deployment-sensitive metric. Here \textbf{CatBoost (E5) leads at 0.8216}, well above E6 (0.7928), E7 (0.7828), and E4 (0.7704); the E5--E6 AUC-PR gap (0.029) is an order of magnitude larger than their AUC-ROC gap (0.001). None of the five LightGBM-routed encoders approach E5's precision-recall performance, and no encoder dominates both metrics. We fixed AUC-ROC as the primary metric in advance to prevent metric-shopping.

\subsection{Robustness and sensitivity}

\textbf{Data scarcity.} We conducted subsampling at 50/25/10\% (105 runs) for the six tree-routed encoders (E1--E6). All six exhibited nearly parallel degradation from 50\% to 10\%, with $-0.031$ to $-0.043$ AUC-ROC loss over that interval, and \textbf{E6 remaining the highest performer at every fraction} (0.9527/0.9391/0.9157 at 50/25/10\%). Over the same 50\%-to-10\% interval, E4 lost slightly more than E6 ($-0.043$ vs.\ $-0.037$), so \textbf{the hypothesis that learned embeddings degrade faster than tier grouping is not supported}. The contrast is driven by E7: off-the-shelf \textbf{TabNet collapsed} to 0.8108 at 10\%, a loss roughly 3$\times$ any tree-routed encoder's, with variance exploding (std $=$ 0.061 vs.\ $\leq$0.004). This finding is scoped to stock TabNet; tuning or regularization may improve results.

\textbf{Per-column localization.} E2, E4, and E6 were evaluated (120 runs) with training on one high-cardinality column at a time. Results indicate \textbf{E6 underperformed classical encoders on all eight columns} ($-0.0009$ to $-0.0045$ AUC-ROC), confirming that E6's edge is attributable to \emph{joint multi-column embedding}: latent co-occurrence across \texttt{card1}, \texttt{addr1}, and \texttt{DeviceInfo}, rather than any single column.

\textbf{Tier-grouping sensitivity.} As shown in Table~\ref{tab:tier}, E4 is largely insensitive to its hyperparameters: AUC-ROC moves less than 0.0019 across $K{=}3\!\to\!7$ for equal-frequency binning, the best sub-variant at every $K$. The $K{=}5$ headline config (0.9548) is virtually equal to the best ($K{=}7$, 0.9556) within one standard deviation.

\begin{table}[!t]
\renewcommand{\arraystretch}{1.0}
\caption{E4 tier-grouping sensitivity (mean AUC-ROC $\pm$ std, $n{=}15$ per cell).}
\label{tab:tier}
\centering
\footnotesize
\begin{tabular}{@{}lccc@{}}
\toprule
Binning rule & $K{=}3$ & $K{=}5$ & $K{=}7$ \\
\midrule
\textbf{equal-frequency} & 0.9537\,$\pm$\,0.0022 & 0.9548\,$\pm$\,0.0025 & \textbf{0.9556\,$\pm$\,0.0021} \\
KS-greedy & 0.9514\,$\pm$\,0.0022 & 0.9515\,$\pm$\,0.0022 & 0.9516\,$\pm$\,0.0021 \\
chi-square & 0.9504\,$\pm$\,0.0023 & 0.9504\,$\pm$\,0.0021 & 0.9506\,$\pm$\,0.0023 \\
\bottomrule
\end{tabular}
\end{table}

\section{Discussion}

\textbf{Encoding as a confounder.} The confounding role of encoding is evident in the work of Grinsztajn et~al.~\cite{grinsztajn2022treebased} and Shwartz-Ziv and Armon~\cite{shwartzziv2022tabular}, who varied the model family along with default encodings. Our approach fixes the learner and varies only the encoder, finding a $\approx$0.01 AUC-ROC swing attributable to encoding alone. We do not claim this matches the tuned, cross-dataset GBDT-vs-DL gaps those authors report, but our structural point stands: encoder choice is an independent degree of freedom that the trees-versus-deep comparisons controlled for implicitly and could not observe.

\textbf{Extracted embeddings versus TabNet.} The two deep-network conditions sit at opposite rank extremes: E7 (target-encoded input to TabNet) is worst and E6 (extracted embeddings to LightGBM) is best. Because E7 shares E2's target encoding, the E7--E2 gap ($-0.0049$, 1579 features each) isolates the downstream learner, showing untuned TabNet trails LightGBM on identical input. Therefore, this E6--E7 gap illustrates a richer encoding of the data, as well as a more capable learner, and whether tuned TabNet, FT-Transformer~\cite{gorishniy2021fttransformer}, or other deep tabular architectures will help close this gap remains open. In addition, task-specific evaluations have the potential to reverse the expected complexity rank order of conditions in other domains. For example, Cao et~al.~\cite{cao2026taskspecific} found that smaller language models perform better than larger models on targeted tasks, which also reflects the result in this study where the most complex pipeline did not outperform all others. A parallel observation holds for representation learning: Lai et~al.~\cite{lai2026transformers} report that transformer-based embeddings do not consistently outperform simpler semantic encoders on short-text classification, echoing our finding that a more elaborate encoder is not universally superior. Using the controlled LightGBM comparisons, we can cleanly isolate E6 compared to E4 ($p{<}0.001$) and E4 compared to E2 ($p{=}0.0003$).

\textbf{Transparency costs.} E4's transparency costs $+0.0064$ AUC-ROC versus E6 and 0.0023 versus E2; both gaps are significant but relatively small. E4 ties E7 ($p{=}0.089$) at 1/67 the cost and trails E2/E5/E6 by at most 0.0064. Under SR~11-7, an auditor can directly inspect tier boundaries (e.g., \texttt{card6}: debit-or-credit at 2.0\% in tier~1, credit at 6.6\% top tier); a 30-dim embedding vector offers no such reading. The interpretability gap that Borisov et~al.~\cite{borisov2022survey} identify can, for the encoding axis, be addressed by tier grouping.

\textbf{Metric sensitivity.} AUC-ROC and AUC-PR disagree on the top performer: E6 leads AUC-ROC (narrowly, tying CatBoost), while E5 is far ahead on AUC-PR (0.8216 vs.\ 0.7928). This lead reflects CatBoost's bundled encoder+learner, not encoding alone, since E5 is a cross-paradigm observation (Section~III). Sun et~al.~\cite{sun2025objective} similarly find, in bank-account fraud, that training objective and class-imbalance handling can matter more than architecture. Our result shows that the encoding decision reshapes rankings independently of the learner. The same pattern recurs across financial modeling, from cross-market volatility forecasting~\cite{cheng2026volatility} to robustness--precision trade-offs in financial retrieval~\cite{cheng2026hybridrag}. The two metrics thus produce different winners, and the choice depends on deployment objectives.

\textbf{Practical recommendation (on IEEE-CIS).} Table~\ref{tab:rec} maps deployment priorities to encoder choice. Maximum accuracy is achieved with E6 and E5; E4 is the preferred choice under model-risk constraints. Off-the-shelf TabNet provides no advantage under this protocol, though a tuned TabNet or alternative deep tabular architecture (e.g., FT-Transformer~\cite{gorishniy2021fttransformer}) could produce different results.

\begin{table}[!t]
\renewcommand{\arraystretch}{1.0}
\caption{Practical recommendation by deployment priority (IEEE-CIS).}
\label{tab:rec}
\centering
\footnotesize
\begin{tabular}{@{}ll@{}}
\toprule
Priority & Recommended encoder \\
\midrule
Highest AUC-ROC & E6 embeddings $\to$ LightGBM \\
Highest AUC-PR & E5 CatBoost (native) \\
Best interpretability--cost & E4 tier grouping \\
Not recommended (untuned) & E7 target enc.$\to$TabNet \\
\bottomrule
\end{tabular}
\end{table}

\section{Conclusion}

On the IEEE-CIS fraud benchmark, a controlled comparison of seven categorical encoding strategies yielded measurable differences in accuracy, computational cost, and interpretability. Learned entity embeddings fed to LightGBM produced the highest AUC-ROC (0.9612), slightly ahead of CatBoost and $+0.0064$ above tier grouping, an edge limited to joint multi-column embedding. The tier grouping method proved competitive and fully auditable, trailing target encoding by only 0.0023 while offering tier boundaries suitable for SR~11-7 review.

The off-the-shelf version of TabNet did not outperform tree-based pipelines and collapsed under data scarcity. On AUC-PR, the more deployment-sensitive metric, CatBoost leads by a wide margin; no single encoder dominated both metrics. Whether tuned deep tabular architectures close the gap remains to be tested.

\textbf{Limitations.} (1)~\emph{Single dataset.} We used IEEE-CIS only; we chose depth over breadth and scope all claims accordingly. (2)~\emph{TabNet not exhaustively tuned.} E7 uses default published parameters; a tuned TabNet, FT-Transformer~\cite{gorishniy2021fttransformer}, or NODE~\cite{popov2020node} could produce different results. (3)~\emph{No temporal split.} We applied stratified random CV; a time-ordered split could change rankings, particularly for target-derived encoders (E2, E4) whose smoothed rates may be inflated by future-leaking folds. (4)~\emph{Anonymized features.} V1--V339 are anonymized, so E4's tier boundaries can be inspected numerically but not mapped to business semantics. (5)~\emph{Fixed downstream model.} LightGBM hyperparameters remain unchanged across E1--E4/E6; per-encoding tuning might reorder classical encoders. (6)~\emph{Small AUC deltas} ($\approx$0.97\,pp spread); the practical signal lies in the compute, interpretability, and data-scarcity axes.

\bibliographystyle{IEEEtran}
\bibliography{references}

\end{document}